\documentclass[runningheads]{llncs}
\usepackage{graphicx}
\usepackage{hyperref}
\usepackage{xcolor}

\makeatletter
\AtBeginDocument{%
  \def\doi#1{\url{https://doi.org/#1}}}
\makeatother

\usepackage{multirow}
\usepackage{amssymb}
\usepackage{tabularx}
\usepackage{array,booktabs}
\newcommand {\otoprule}{\midrule [\heavyrulewidth]}
\newcolumntype {+}{ >{\global\let\currentrowstyle\relax}}
\newcolumntype {^}{ >{\currentrowstyle }}
 \newcommand {\rowstyle}[1]{\gdef\currentrowstyle{#1} %
 #1\ignorespaces
 }
\newcommand{\tabhead}{\rowstyle{\bfseries}}
\usepackage{xspace}
\newcommand{\basern}{ResNet\xspace}
\newcommand{\basecn}{ConvNeXt\xspace}
\newcommand{\pip}{PIPNet\xspace}
\newcommand{\piprn}{PIPNet$_{RN}$\xspace}
\newcommand{\pipcn}{PIPNet$_{CN}$\xspace}
\newcommand{\pipek}{PIPNet$_{EK}$}

\usepackage{xr}
\begin{document}

\title{PIPNet3D: Interpretable Detection of Alzheimer in MRI Scans}

\titlerunning{PIPNet3D: Interpretable Detection of Alzheimer}

\author{
Lisa Anita De Santi\inst{1,6}, 
Jörg Schlötterer\inst{2,3}, 
Michael Scheschenja\inst{4}, 
Joel Wessendorf\inst{4}, 
Meike Nauta\inst{5}, 
Vincenzo Positano\inst{6}, 
Christin Seifert\inst{2}
}
\authorrunning{Anonymous}

\institute{
University of Pisa, Pisa, Italy \and 
University of Marburg, Marburg, Germany \and 
University of Mannheim, Mannheim, Germany \and
Department of Diagnostic and Interventional Radiology, University Hospital Marburg, Marburg, Germany \and
Datacation, Eindhoven, Netherlands \and 
Fondazione Toscana G Monasterio, Pisa, Italy
}
\maketitle              
\begin{abstract}
Information from neuroimaging examinations is increasingly used to support diagnoses of dementia, e.g., Alzheimer's disease.
While current clinical practice is mainly based on visual inspection and feature engineering, Deep Learning approaches can be used to automate the analysis and to discover new image-biomarkers. Part-prototype neural networks (PP-NN) are an alternative to standard blackbox models, and have shown promising results in general computer vision. 
PP-NN's base their reasoning on prototypical image regions that are learned fully unsupervised, and combined with a simple-to-understand decision layer. 
We present PIPNet3D, a PP-NN  for volumetric images. 
We apply PIPNet3D to the clinical diagnosis of Alzheimer’s Disease from structural Magnetic Resonance Imaging (sMRI).
We assess the quality of prototypes under a systematic evaluation framework, propose new functionally grounded metrics to evaluate brain prototypes and develop an evaluation scheme to assess their coherency with domain experts. 
Our results show that PIPNet3D is an interpretable, compact model for Alzheimer’s diagnosis with its reasoning well aligned to medical domain knowledge. Notably, PIPNet3D achieves the same accuracy as its blackbox counterpart; and removing the remaining clinically irrelevant prototypes from its decision process does not decrease predictive performance.


\end{abstract}

\section{Introduction}

Alzheimer’s Disease (AD) is a neurodegenerative disease resulting in a progressive decline of cognitive abilities. 
The diagnosis of AD is typically integrated with neuroimaging examinations, e.g., structural Magnetic Resonance Imaging (sMRI)~\cite{DeSanti2023,Chandra2019,Vemuri2010}. 
Dementia patients exhibited pathological patterns in sMRI, such as gray matter atrophy and other tissue abnormalities in specific brain areas~\cite{Chandra2019}. 
Such existing quantitative tools \cite{Fischl2012} present limited detection capabilities ~\cite{Chandra2019,Vemuri2010}, Deep Learning (DL) models can facilitate the analysis of sMRI and have the potential to extract yet unknown image-based biomarkers in AD. 
However, the blackbox nature of DL models makes their application in high-stakes decisions controversial.
Explainable Artificial Intelligence (XAI) aims to develop transparent systems while maintaining the advantages of DL~\cite{VanDerVelden2022,Rudin2019,Salahuddin2022}.
For the application of XAI models in real-world scenarios, objectively assessing the quality of explanations is crucial~\cite{Longo2024}. 
However, evaluating explanations is considered challenging, such there is no ground truth of ``good explanations''~\cite{Salahuddin2022} and due to their multidisciplinary nature ~\cite{Miller2019}. 
In addition, previous studies highlighted the centrality of involving domain experts in the development stage of XAI clinical supporting-decision system ~\cite{Cabitza2023}, but most of the current XAI works did not include a human evaluation approach~\cite{Longo2024}.
As posthoc XAI approaches may not faithfully reflect a model's reasoning~\cite{Rudin2019}, there is growing interest in self-explanatory models, where the model itself inherently provides the explanation. One such type of models are part-prototype neural networks (PP-NN), which base their reasoning on the detection of machine-learned \textit{prototypical parts}~\cite{Chen2018}.
Several works applied DL for AD diagnosis~\cite{Khojaste2022} from sMRI. While the majority use blackbox models, a few also applied PP-NN~\cite{Mohammadjafari2021,Mulyadi2023}.
However, their models can not consider spatial relationships in the 3rd dimension~\cite{Mohammadjafari2021}, or use non-interpretable (nonlinear) decision layers~\cite{Mulyadi2023}.
There is an increasing interest in extending PP-NN models from general computer vision tasks to 3D medical images~\cite{Wolf2023,Wei2023}.
However, those works are based on ProtoPNet~\cite{Chen2018}, which has been shown to lack in compactness of the explanation~\cite{Rymarczyk2022} and semantic quality of prototypes~\cite{Nauta2023_PIPNet}.
Among PP-NN models, PIPNet reported appealing properties for the medical domain~\cite{Nauta2023_PIPNet_MI}, because it learns a small number of semantically meaningful prototypes, enables human interaction and direct manual adaptation of the model's reasoning, and detects out-of-distribution data, abstaining from decisions when there is not sufficient evidence for any of the classes. 
As PIP-Net was designed for 2D images, and processing 3D volumes with a 2D backbone might induce information loss, we introduce PIPNet3D: an interpretable classifier able to process 3D scans. 

In summary, our contributions are the following: 
(1) We introduce PIPNet3D, an interpretable part-prototype classifier for 3D input data, based on the \pip 2D model~\cite{Nauta2023_PIPNet}. 
(2) We apply PIPNet3D to AD diagnosis from sMRI, and find that PIPNet3D performs equally well to its corresponding blackbox baseline.
(3) We thoroughly assess PIPNet3D w.r.t. comprehensive explanation desiderata and propose \textit{Prototype Brain Entropy} and \textit{Prototype Localization Consistency} as novel measures for functionally grounded evaluations. 
(4) We proposed a domain experts' evaluation setup generalizable for the assessment of other PP-NN architectures.
(5) From the domain experts' evaluation we find that i) PIPNet3D aligns well with domain knowledge and ii) removing clinically irrelevant prototypes improved the model's compactness without impacting performances. 
Source code and trained models are available at \url{https://anonymous.4open.science/r/PIPNet3D-58CA}.

\section{Methodology}
This section introduces the architecture of PIPNet3D (cf. Sect.\ref{sec:PIPNet3D} and Fig.~\ref{fig:PIPNet3D}), which enables an understandable decision structure of the model (cf. Fig.~\ref{fig:explanation}). 
We describe the evaluation, including the novel metrics \textit{Prototype Brain Entropy} and the \textit{Prototypes Localization Consistency}, and expert evaluation in Sect.~\ref{sec:proto_eval}. 

\subsection{PIPNet3D}
\label{sec:PIPNet3D}

\begin{figure}[bt]
\centering
\includegraphics[width=0.75\textwidth]{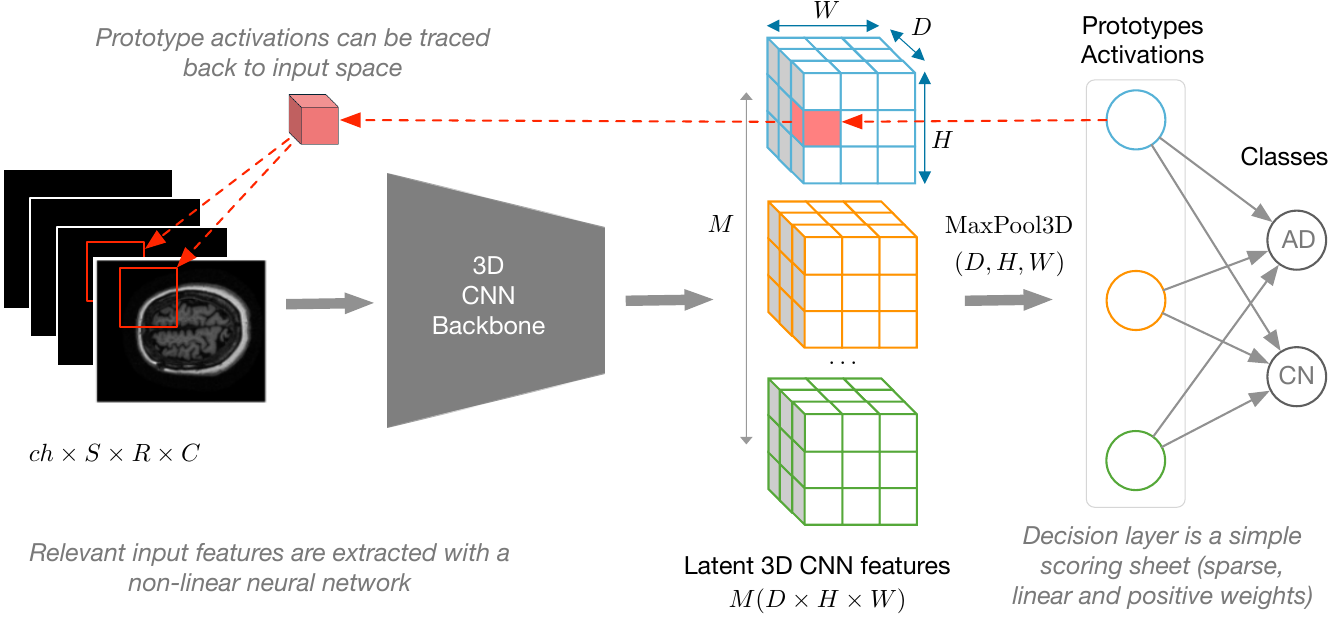}
\caption{Overview of PiPNet3D. 3D prototypes are learned through a CNN backbone. Representations are optimized through a contrastive pre-training step. A linear sparse decision layer computes the predictions based on prototype (VOI) activations.} 
\label{fig:PIPNet3D}
\end{figure}

\begin{figure}[t]
\centering
\includegraphics[width=0.75\textwidth]{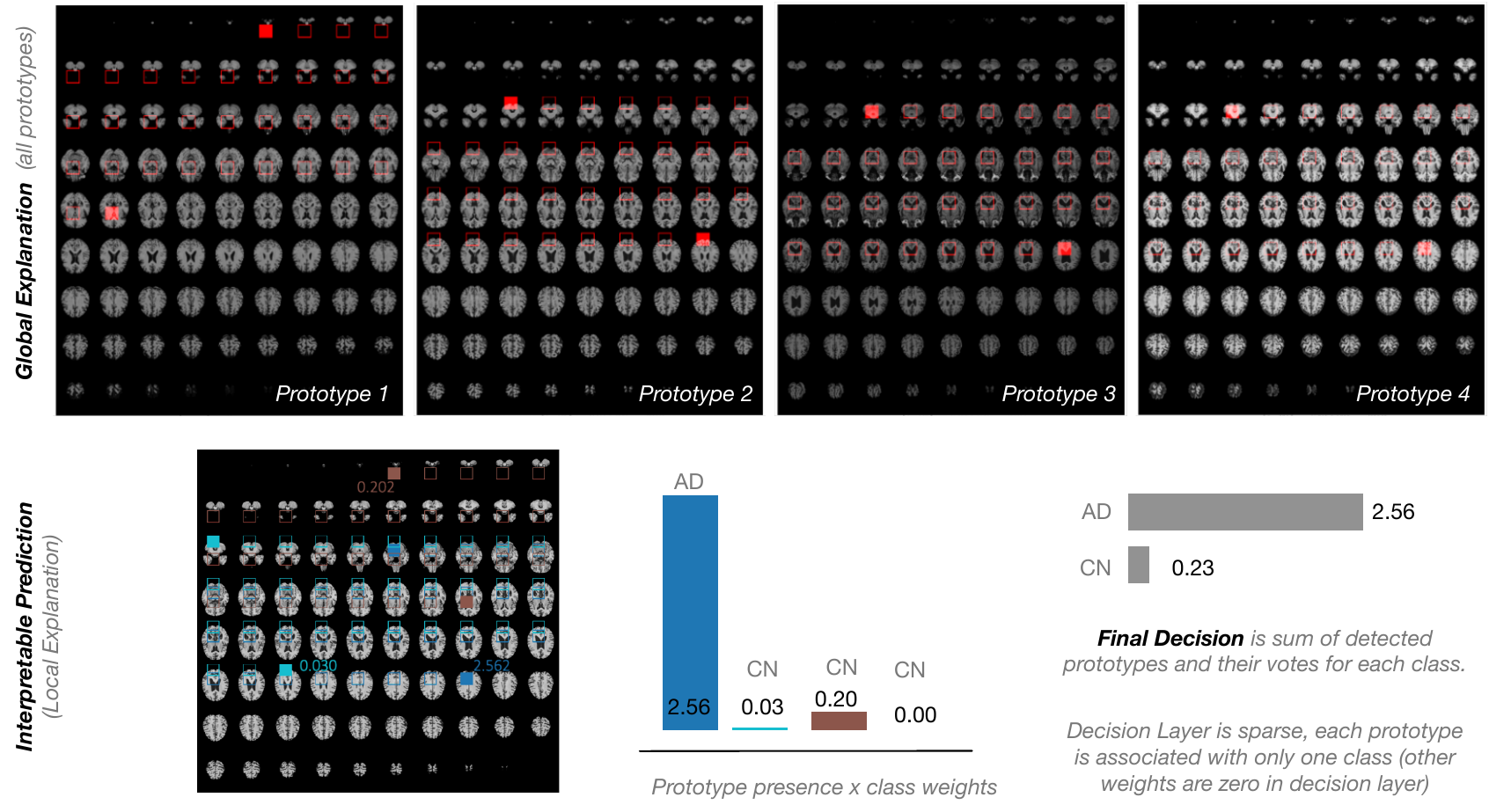}
\caption{Example of local and global explanation (LE and GE). The GE shows all the learned prototypes. The LE shows the model's reasoning for one particular patient. } 
\label{fig:explanation}
\end{figure}

We designed our PIPNet3D based on a 3D-CNN feature extractor and follow the scoring-sheet reasoning and training paradigm of Nauta et al.~\cite{Nauta2023_PIPNet}. The architecture (cf. Fig.~\ref{fig:PIPNet3D}) consists of an input layer, a CNN backbone, global max pooling of the feature maps, and a linear classification layer. 
The \textbf{input layer} is of dimension $\mathbf{x}\in\mathbf{R}^{ch\times S\times R\times C}$ where $ch,S,R,C$ respectively represents the number of channels, slices, rows and columns which constitutes the input volume.
The \textbf{CNN backbone}, $z=f(\mathbf{x})$ extracts the latent features consisting of $M$ 3-dim $(D\times H\times W)$ feature maps where $z_{m,d,h,w}$ represents to the one-hot encoding of patch $d,h,w$ to the prototype $m$.
A \textbf{global max-pooling 3D} operation extracts $M$ prototypes and calculates the prototypes presence scores $\mathbf{p}$ where $p \in [0., 1.]^M$ and $p_j$ measures the presence of the prototype $m$ in the input image.
The \textbf{linear classification layer} of size $\mathbf{w_c}\in\mathbf{R}^{M\times K}_{\geq 0}$ connects prototypes to the classes acting as a scoring sheet, where $w_c^{m,k}$ represents the contribution of prototype $m$ for class $k$. In our use case $k=2$ classes: AD (Alzheimer's disease) and CN (Cognitively Normal). This layer is optimized to be sparse (i.e., having as few connections as possible). The final output is a score for each of the  $K$ classes, where every \textbf{class output score} is given by the sum of the prototypes' presence score weighted by the relevance of each prototype to that class: $\mathbf{o}=\mathbf{p}\cdot\mathbf{w_c}$, where $\mathbf{o}$ is $1\times K$ and $o_k=\sum_{d=1}^D p_dw_c^{d,k}$.
As CNN backbones we used ResNet-18 3D pretrained on Kinetics400~\cite{Tran2017} and ConvNeXt-Tiny 3D pretrained on the  STOIC medical dataset~\cite{Kienzle2022}, denote respectively as \piprn and \pipcn.
This results in a number of initial prototypes $M=512$ and $M=768$.
We applied data augmentation that does not alter the information content relevant to the clinical task~\cite{Khojaste2022} and finetuned PIPNet3D re-using the hyperparameter settings of the original PIPNet~\cite{Nauta2023_PIPNet}. We only changed the batch size $=12$ to adapt it to our computational capabilities.
We reported the details on the data augmentation pipeline and hyperparameter setting in Suppl. Material.
After training, we selected the relevant prototypes, i.e., those that have a weight connection $w_c>10^{-3}$ for at least one class and were detected with a similarity $p>0.1$ in the training set~\cite{Nauta2023_PIPNet}.
The global explanation (cf. Fig.~\ref{fig:explanation}) consists of relevant prototypes of the most similar images in the training set marked as a volume of interest (VOI).
During inference, PIPNet3D returns the predicted class and the annotation of every prototype detected with similarity $p>0.1$, together with its similarity score. 

\subsection{Neuroimaging Prototypes Explanation Quality}
\label{sec:proto_eval}
We assessed the explainability using the Co-12 evaluation framework~\cite{Nauta2023_Co12,Nauta2023_Co12_PP} (details in Suppl. Material). 
We designed an automated quantitative functionally-grounded evaluation setup, and an evaluation setup with domain experts. 
For both setups, we defined quantitative metrics to perform objective measurements relevant to the neuroimaging domain.
We selected the PIPNet3D backbone which obtained the highest predictive performances and applied the evaluation to compare the quality of the prototypes from the 5 different folds (denoted as \textbf{Mx} where \textbf{x} indicates the current fold) and to study their relationship with their diagnostic performance.

\subsubsection{Functionally Grounded Evaluation.}
We measured the compactness of PIPNet3D by calculating the total number of prototypes (\textit{Global size}), the average number of detected prototypes in a test image (\textit{Local size)}, and the \textit{Sparsity} of the decision layer following previous work~\cite{Nauta2023_PIPNet_MI}.
Additionally, we propose two quality measures for prototypes based on brain reference standards to anatomically localize and study the composition of the extracted prototypes.

We propose \textit{Prototype Brain Entropy} $H_{p}$ to measure prototype purity.
We used the Cerebrum Atlas (CerebrA) corresponding to the ICBM2009c space to annotate the prototypes with the anatomical brain regions~\cite{Manera2020}.
For every prototype $p_i$ we selected the corresponding $VOI_{p}$ in the atlas, counted the voxels of every brain region, and computed the Shannon Entropy $H(.)$ over brain regions. 
A ``pure'' prototype focused on one specific brain area has an $H_{p}=0$, while a higher value indicates the inclusion of multiple regions. An explanatory Figure is reported in Suppl. Material.
\begin{equation}
    H_{p} = H(CerebrA_{VOI_{p}})
\end{equation}

We propose \textit{Prototypes Localization Consistency} to measure whether the same prototype is activated in similar brain regions. 
We generate the local explanations for every image $img$ in the test set, and evaluate the coordinate center ($cc$) of the VOI representing each prototype $p$, $VOI_{cc,p}|_{img}$.
Next, we compute the average prototype's $cc$ $\overline{VOI}_{cc,p}$. 
We then compute the Euclidean distance between $VOI_{cc,p}|_{img}$ and $\overline{VOI}_{cc,p}$ and normalize for the maximum linear dimension of the $VOI$ (for a cubic VOI $l\sqrt{3}$, where $l$ is the side of the cube).
We define \textit{Prototypes Localization Consistency}, $LC_{p}$ of $p$ by averaging over the test set.
$LC_{p}$ ranges between $0-1$ where $0$ expresses the maximum localization consistency of the prototype.
A prototype consistently located ($LC_{p} \approx 0$) can be associated with a specific brain region in the ICBM2009c standard space.
\begin{equation}
    LC_{p} = \sum_{img}\frac{||VOI_{cc_{p}}|_{img}-\overline{VOI}_{cc}||^2}{l\sqrt{3}} 
\end{equation}

\subsubsection{Evaluation with Experts}
\label{sec:exp_eval}
We collected quantitative and qualitative feedback on the prototypes from two radiologists at the University Hospital of Marburg. 
Our survey prompted the participants to assess the following statements.\footnote{We performed two pre-tests, and only report on the final survey design.}
``The prototype is located in a brain region typically affected in AD.'' (\textit{Localization coherence}). ``A prototype adding evidence to CN does not show pathological patterns, a prototype adding evidence to AD shows an abnormal pattern.'' (\textit{Pattern coherence}). ``The prediction returned by the model based on the VOI is correct.'' (\textit{Classification coherence}). For each item we used a 6-point Likert scale (1: strongly disagree -- 6: strongly agree).
Additionally, we asked participants ``Which clues in this VOI vote for or against each diagnosis?'' (free text).
The radiologists assessed 35 prototypes in total. Each prototype was shown as i) a montage of the most activated image in the training set with the prototype marked with a VOI; ii) a detailed view montage of the extracted VOI.
The survey proposed is an evaluation scheme that can be used to empirically assess the consistency of the prototype with respect to domain knowledge of other PP-NN architectures. An explanatory Figure is reported in Suppl. Material.
We assessed the inter-user agreement for each item using the Interclass Correlation Coefficient (ICC)~\cite{Koo2016}. 
We averaged the scores of the two radiologists for every prototype and binarized the scores considering a prototype with a score $\geq$ 3.5 as coherent. 
We finally evaluated the rate of coherent prototypes. 
Additionally, we tested a human-in-the-loop setting by suppressing the incoherent prototypes from the model (setting \pipek in Table~\ref{tab:results:overall}).

\section{Evaluation}
Data used in the preparation of this article were obtained from the Alzheimer’s Disease Neuroimaging Initiative (ADNI) database\footnote{\url{https://adni.loni.usc.edu}}. The ADNI was launched in 2003 as a public-private partnership, led by Principal Investigator Michael W. Weiner, MD. The primary goal of ADNI has been to test whether serial magnetic resonance imaging (MRI), positron emission tomography (PET), other biological markers, and clinical and neuropsychological assessment can be combined to measure the progression of mild cognitive impairment (MCI) and early Alzheimer’s disease (AD).
Selecting the \textit{``ADNI1 Standardized Screening Data Collection for 1.5T scans''} processed with Gradwarp, B1 non-uniformity, and N3 correction, we obtained 307 CN and 243 AD sMRI brain scans.
Our data pre-processing pipeline is inspired by Mulyadi et al.~\cite{Mulyadi2023}, detailed in Suppl. Material.
We performed 5-fold cross-validation with patient-wise random splits. $20\%$ of training images are used for validation. 
We compared \piprn and \pipcn to their corresponding blackbox ResNet-18 and ConvNeXt-Tiny 3D models, which we fine-tuned with dynamic data augmentation using the same augmentation pipeline of \pip. Prediction performances were compared using paired t-test.
We report hyperparameter settings in Suppl. Material.
We implemented PIPNet3D using PyTorch and MONAI\footnote{{\url{https://monai.io}}}. 
We trained our models on an Intel Core i7 5.1 MHz PC, 32 GB RAM, equipped with an NVIDIA RTX3090 GPU with 24 GB of embedded RAM. The overall training process took on average 1 hour and 25 minutes for PIPNet3D and 1 hour for the blackbox.

\subsection{Predictive Performance}
We evaluated the classification performances using Balanced Accuracy, Specificity, Sensitivity and F1 (cf. Table~\ref{tab:results:overall}).
The ResNet-18 model obtained the highest predictive performances both as blackbox and \pip backbone.
Paired two-tailed t-tests, $\alpha=0.05$ showed no significant differences in performance between all 5 folds, suggesting that PIPNet3D adds interpretability to the corresponding blackbox model without reducing classification performance.
We report classification results and interpretability of related works (not directly comparable to ours) in Suppl. Material.

\begin{table}[tb]
\centering
\caption{Performances (5 folds, in percentages) of the black-box models, and the interpretable \pip. \pipek are models after pruning prototypes that do not align with expert knowledge (cf. Sect.\ref{sec:exp_eval}). $\varnothing$: averaged over 5 folds, *: best.}
\label{tab:results:overall}
\setlength{\tabcolsep}{6pt}
\begin{tabular}{+l^c^c^c^c}
\toprule \tabhead
            & Balanced Accuracy  & Specificity & Sensitivity & F1 \\
\otoprule
\basern     & 80 $\pm$ 06 & 78 $\pm$ 15 & 82 $\pm$ 12 & 79 $\pm$ 07\\
\piprn      & 82 $\pm$ 02 & 88 $\pm$ 07 & 76 $\pm$ 09 & 79 $\pm$ 03\\[4pt]
\basecn     & 61 $\pm$ 07 & 67 $\pm$ 15 & 56 $\pm$ 24 & 54 $\pm$ 15\\
\pipcn      & 69 $\pm$ 03 & 71 $\pm$ 07 & 68 $\pm$ 08 & 66 $\pm$ 04\\[4pt]
\midrule
\multicolumn{5}{c}{\textsc{After aligning with expert knowledge}} \\[2pt]
\pipek$^\varnothing$ & \textbf{82} $\pm$ 02 & 88 $\pm$ 07 & 74 $\pm$ 12 & 78 $\pm$ 05 \\
\pipek$^{*}$ & 85	      & 84	         & 86	       & 83 \\
\bottomrule
\end{tabular}
\end{table}

\subsection{Interpretability}
Table~\ref{tab:explanation_quality} reports on the quality of prototypes.
All folds of PIPNet3D have a small number of prototypes (Global size, max 11), while initially starting with a setup of 512 prototypes for \piprn. Additionally, we observed that prototypes can uniquely be assigned to one class (e.g., the connection to the other class is zeroed out), leading to compact and comprehensible models.
Most of the prototypes are consistently located in the same anatomical location (small $H_{p_i}$, and $LC_{p_i}$).
While all our models have a relatively small size (average 2.4 to 5.4 prototypes in local explanations), a Pearson correlation of 0.91 ($\pm$0.03) between F1 score and Local size suggest that models with high accuracy are those with less compact local explanations (more prototypes). 
A Pearson correlation of -0.89($\pm$0.04) between F1 and Prototype Entropy $H_p$, shows that a model with purer prototypes also has higher accuracy.
Our evaluation with domain experts showed a good reliability (ICC: 0.80, 0.76 and 0.85 for the quantitative tasks)~\cite{Koo2016}. 
Prototypes were found to be coherently located (coherence score 0.70, cf. Table~\ref{tab:explanation_quality}), exhibit a coherent pattern (score 0.90), and classification decision (score 0.90).
Most prototypes incoherently located are associated with the CN class (Localization coherence of 0.54, AD: 0.87). Incoherent CN prototypes might be due to PiPNet3D training scheme that requires finding evidence for every class; even if this class is representing absence of abnormalities. 
We observed that most of the prototypes showed a coherent pattern with the connected class. This indicates that most of the prototypes provide evidence coherent with the medical knowledge to support the model's decision. In this case, we observed a lower pattern coherence for the AD prototypes (0.77 vs. 0.93). We also noticed that most of the AD incoherent prototypes were observed in M3 and M4, where the model reported a lower recall for the AD class.
We further analyzed the qualitative feedback for the prototypes that did not show a coherent pattern.  For one CN prototype, both radiologists reported the presence of dilated ventricles as a viable sign of AD, suggesting incorrect model reasoning. For two AD prototypes, the VOI showed no or only mild pathological clues. This may again point to incorrect model reasoning or indicate yet unknown patterns, that may not be visible to a human observer.
We suppressed the prototypes that did not show a coherent pattern and we did not observe statistically significant differences in classification performance in the adapted models w.r.t. the original ones (cf. Table~\ref{tab:results:overall}, models \pipek). Thus, adapting to domain knowledge increased the models' compactness and coherency, without negatively impacting their accuracy.
\begin{table}[tb]
\caption{Evaluation of prototype quality for models M1, ..., M5. For the evaluation with experts, we show median rating in brackets.  $\uparrow$ and $\downarrow$: tendency for better values.}
\label{tab:explanation_quality}
\setlength{\tabcolsep}{4pt}
\begin{tabular}{+l^l^c^c^c^c^c^c}
\toprule \tabhead
&                   & M1    & M2    & M3    & M4    & M5    & Average \\\otoprule
\multirow{11}{*}{\rotatebox[origin=c]{90}{\textsc{Functional Evaluation}}}
&Global size $\downarrow$     & 10    & 11    & 5     & 5     & 4     & 7.0\\
&~~~~ for CN     & 5     & 4     & 3     & 3     & 3     & 3.6\\
&~~~~ for AD     & 5     & 7     & 2     & 2     & 1     & 3.4\\

&Local size $\downarrow$     & 5.4   & 5.2   & 2.7   & 3.0   & 2.4   & 3.8\\

&Sparsity $\uparrow$       & 0.990  & 0.989  & 0.995 & 0.995 & 0.996 & 0.993 \\

&$LC_p$ $\downarrow$	& 0.004 & 0.021 & 0.008 & 0.018 & 0.030 & 0.016 \\
&~~~~ for CN     & 0.009  & 0.003  & 0.000  & 0.015  & 0.030  & 0.017 \\
&~~~~ for AD     & 0.000  & 0.016  & 0.020  & 0.022  & 0.050  & 0.022 \\

&$H_p$ $\downarrow$	& 2.5   & 3.1   & 3.4  & 3.1   & 3.4   & 3.1\\ 
&~~~~ for CN     & 2.8   & 3.3   & 3.5  & 3.1   & 2.9   & 3.1\\
&~~~~ for AD     & 2.3   & 2.9   & 3.3  & 3.1   & 3.8   & 3.1\\\midrule

\multirow{9}{*}{\rotatebox[origin=c]{90}{\textsc{Users}}}
&Localization Coherence $\uparrow$ & 0.90   & 0.60   & 0.60   & 0.80   &0.50    & 0.70 (3.5)\\
&~~~~ for CN     & 0.80  & 0.25  & 0.67  & 0.67  & 0.33  & 0.54 \\
&~~~~ for AD     & 1.00  & 0.86  & 0.50  & 1.00  & 1.00  & 0.87 \\
                
& Pattern Coherence $\uparrow$%
                & 1.00   & 0.90   & 0.80   & 0.80   & 0.80    & 0.90 (4.5)\\
&~~~~ for CN     & 1.00  & 1.00  & 1.00  & 1.00  & 0.67  & 0.93\\
&~~~~ for AD     & 1.00  & 0.86  & 0.50  & 0.50  & 1.00  & 0.77\\

&Classification Coherence $\uparrow$
                & 1.00   & 0.90   & 0.80   & 1.00   &0.80    & 0.90 (4.5)\\
&~~~~ for CN     & 1.00  & 1.00  & 1.00  & 1.00  & 0.67  & 0.93 \\
&~~~~ for AD     & 1.00  & 0.86  & 0.50  & 1.00  & 1.00  & 0.87\\\bottomrule
\end{tabular}
\end{table}

\section{Conclusion}
In summary, we introduced PIPNet3D, a part-prototype model for 3D input images, applied to diagnose AD from sMRI, whose interpretability has been systematically assessed.
We obtained an interpretable and compact model able to support AD diagnosis with prototypes in line with medical knowledge. 
Our results show, that interpretable models that align with domain-knowledge can be as effective as blackbox models. We hope to inspire their usage in the future, especially in high-risk domains, such as medical image analysis. 
We plan to extend PIPNet3D to a multi-modal system additionally taking patients' demographic information, introducing intermediate levels of cognitive impairments, and evaluating OoD performances with intermediate levels of dementia and patients with different age groups.

%
\clearpage
\bibliographystyle{splncs04}
\bibliography{mybibliography}

\end{document}


\title{PIPNet3D: Interpretable Detection of Alzheimer in MRI Scans}
\author{}
\date{}
\maketitle    

\textbf{Image Preprocessing}: (1) Image registration to the ICBM152 Non-Linear Symmetric 2009c standard space \cite{Fonov2009} with affine registration using SimpleITK \cite{Yaniv2018}. (2) Gray matter selection with the ICBM152 Non-Linear Symmetric 2009c brain mask. (3) Empty slices removal along the transverse axes (margin $=3$ to the brain mask), $2$ downsampling (reduce data dimensionality and speed up computation), final image size: $80\times 114\times 96$. (4) Min-max Intensity normalization to the range $[0-1]$.
 
\textbf{Data Augmentation}: Random rotation $\pm 10^{\circ}$ along x, y, z; random shift $\pm 5$ along x, y, z; random zoom $0.9–1.1$ and random Gaussian noise $N(0,0.01)$. We finally repeated the channel to $3$ to match the ResNet18-3D expected input dimension.

\begin{table}[H]
\captionsetup{skip=0pt}
\centering
\caption{\textbf{Hyperparameter settings}. PIPNet3D trained re-using the hyperparameters setting from the original paper~\cite{Nauta2023_PIPNet} and adapted batch size to our computational capabilities. ResNet18-3D trained re-using hyperparameters setting from~\cite{Mulyadi2023} which implemented the ResNet18-3D architecture in the work) and adapted batch size (same batch size of PIPNet3D)}
\label{tab:hyp}
\small
\begin{tabularx}{\textwidth}{+X^X}
\toprule\tabhead
PIPNet3D & ResNet18-3D  \\ \midrule
(1) Self-Supervised learning of prototypes: Adam optimizer, backbone learning rate $0.0001$, linear classification learning rate $0$ (frozen), batch size $12$, $10$ epochs; \newline
(2) PIPNet training: Adam optimizer, backbone learning rate $0.0001$, linear classification learning rate $0.05$, batch size $12$, $60$ epochs.
& 
Adam optimizer, learning rate $10^{-4}$, batch size $24$, $100$ epochs . 
 \\ \bottomrule
\end{tabularx}
\end{table}

\begin{table}[H]
\captionsetup{skip=0pt}
\centering
\caption{Classification performances (F1) and model's interpretability of most relevant \textbf{related work} in Alzheimer's diagnosis}
\label{tab:literature}
\small
\begin{tabularx}{\textwidth}{+l^l^X}
\toprule\tabhead
Authors & F1 & Models and Interpretability  \\ \otoprule
\cite{Jin2023} & 0.88 & ResNet18-3D with Attention Mechanism producing explanation heatmaps  \\ 
\cite{Korolev2017} & 0.88 & Adapted 3D VoxCNN and ResNet with posthoc explanation heatmaps   \\ 
\cite{Liu2022} & 0.89 & 3D CNN with posthoc explanation heatmaps and t-SNE \\ 
\cite{Mulyadi2023} & \textbf{0.91} & ResNet18-3D with Prototypical Latent Space classified by a NN  \\ \midrule
Ours & 0.79 & 3D ResNet-18 baseline black-box model  \\ 
~ & 0.79 & PiPNet3D: Interpretable-by-design PP model \\ \bottomrule
\end{tabularx}
\end{table}

\begin{table}[H]
\setlength{\tabcolsep}{4pt}
\captionsetup{skip=0pt}
\caption{\textbf{The Co-12 assessment of \pip's interpretability}}
\label{tab:appdx:Co-12-overview}
\small
\begin{tabularx}{\textwidth}{+l^l^X}
\toprule\tabhead
& Criterion & \\\otoprule
\multirow{10}{*}{\rotatebox[origin=c]{90}{\scriptsize\textsc{Content}}}
& Correctness &  
    Classification correct by design. Visualization with upsampling~\cite{Chen2018} might not be fully correct~\cite{Gautam2023}, and is direction of future work.\\
& Completeness &
    Complete by design; the full model behavior (linear layer connections is explained). Representation learning (backbone) as blackbox with interpretable output (prototypes). \\
& Consistency &
    Consistent by design (no random components in the forward pass computations), but may be subject to standard numerical computation differences. We assessed the difference in prototype's localization when detected in different subjects (cf. $LC_p$).\\
 & Continuity &
    Optimized by the contrastive learning setup of \pip, but not explicitly evaluated. \\
 & Contrastivity &  
     Output-sensitivity implemented by-design. 
     \\
 & Cov. complexity &  
      Prototype purity assessed with anatomical brain atlas annotation (cf. $H_p$).\\\midrule
\multirow{4}{*}{\rotatebox[origin=c]{90}{\scriptsize\textsc{Presentation}}}
& Compactness &  
    Evaluated by \textit{global} and \textit{local explanation size.} \\
& Composition &  
    Showing both, overview and detail; localization of the found prototype in the brain, and its detail view. \\
& Confidence &  
    Diagnosis prediction reported together with the prototypes’ similarity score which constitutes the Alzheimer's fingerprint of the subject and the OoD detection confidence. \\\midrule
\multirow{6}{*}{\rotatebox[origin=c]{90}{\scriptsize\textsc{User}}}
& Context &  
 We evaluated prototypes with domain experts in a human-grounded evaluation. \\
& Coherence & 
 User study with radiologists evaluating whether the prototypes align with experts' visual assessment w.r.t. their localization, detected pattern, and diagnostic decision (cf. \textit{Localization Coherence, Pattern Coherence, Classification Coherence}). \\
& Controllability &  
 We use results collected from the coherence analysis to suppress the prototypes that are not in line with the expert evaluation. \\\bottomrule
\end{tabularx}
\end{table}

\begin{figure}[ht]
\captionsetup{skip=0pt}
\centering
\includegraphics[width=0.8\textwidth]{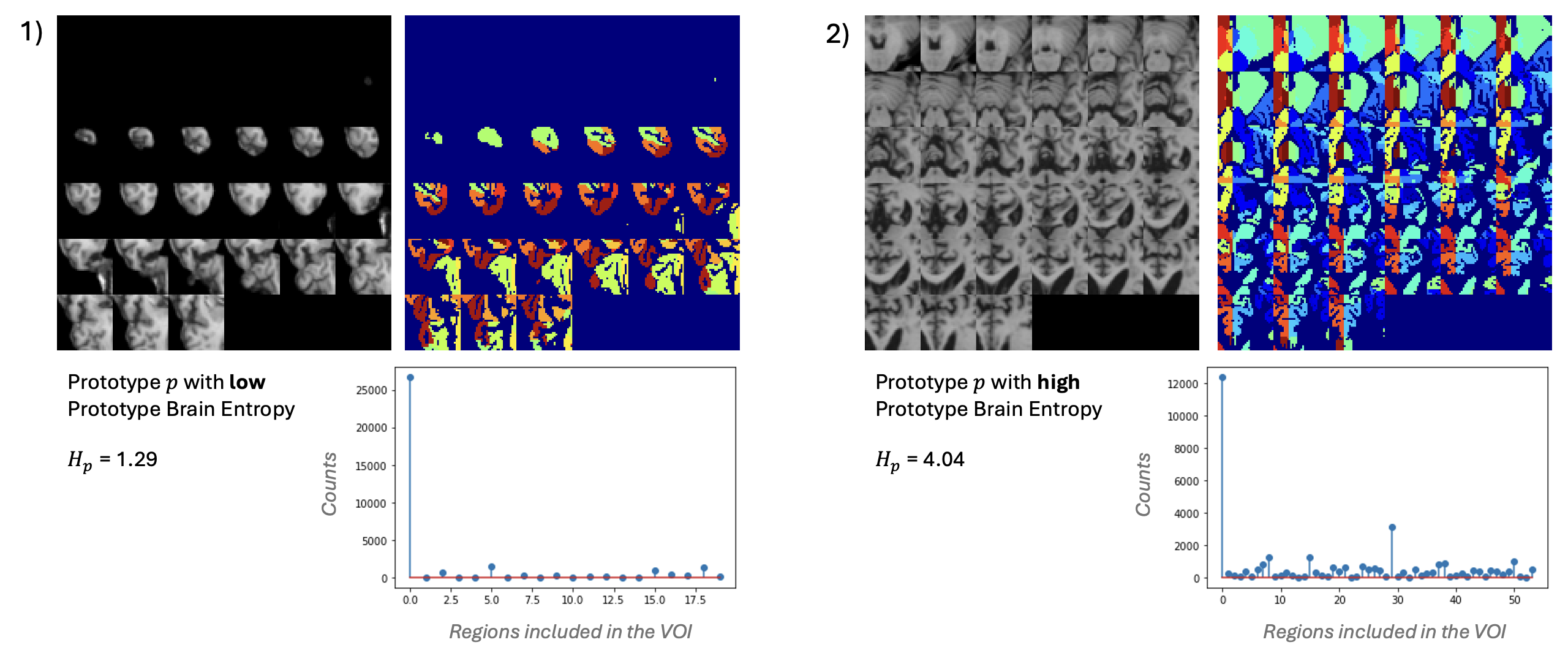}
\caption{Montage of the image VOI, VOI in the CerebrA atlas, and counts of the CerebrA regions in the VOI for two different prototypes. 1) represent a ``pure'' prototype (lower $H_{p}$); 2) represent an ``impure'' prototype (higher $H_{p}$)} 
\label{fig:PIPNet3D}
\end{figure}

\begin{figure}[ht]
\captionsetup{skip=0pt}
\centering
\includegraphics[width=0.8\textwidth]{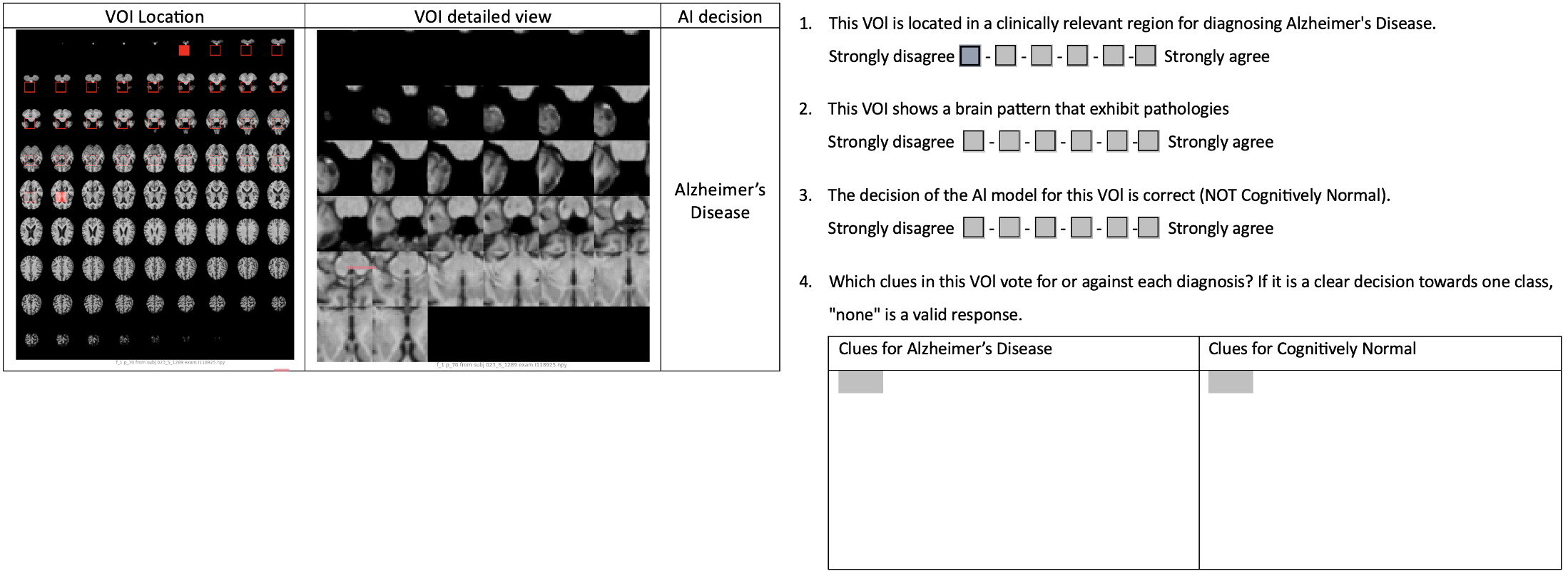}
\caption{Example of the survey used to evaluate prototypes with domain experts} 
\label{fig:survey}
\end{figure}

\bibliographystyle{apalike}
\bibliography{mybibliography}